\documentclass{article}

\usepackage{arxiv}

\usepackage[utf8]{inputenc} % allow utf-8 input
\usepackage[T1]{fontenc}    % use 8-bit T1 fonts
\usepackage{hyperref}       % hyperlinks
\usepackage{url}            % simple URL typesetting
\usepackage{booktabs}       % professional-quality tables
\usepackage{amsfonts}       % blackboard math symbols
\usepackage{nicefrac}       % compact symbols for 1/2, etc.
\usepackage{microtype}      % microtypography
\usepackage{lipsum}
\usepackage{graphicx}
\graphicspath{ {./images/} }

\title{Improvement in land cover and crop classification based on temporal features learning from Sentinel-2 data using Recurrent-Convolutional Neural Network (R-CNN)}

\author{
 Vittorio Mazzia \\
  Department of Electronics and Telecommunications\\
  Politecnico di Torino\\
  10124 Turin, Italy \\
  \texttt{vittorio.mazzia@polito.it} \\
   \And
 Aleem Khaliq \\
  Department of Electronics and Telecommunications\\
  Politecnico di Torino\\
  10124 Turin, Italy \\
  \texttt{aleem.khaliq@polito.it} \\
  \And
 Marcello Chiaberge \\
  Department of Electronics and Telecommunications\\
  Politecnico di Torino\\
  10124 Turin, Italy \\
  \texttt{marcello.chiaberge@polito.it} \\
}

\begin{document}
\maketitle
\begin{abstract}
Understanding the use of current land cover, along with monitoring change over time, is vital for agronomists and agricultural agencies responsible for land management. The increasing spatial and temporal resolution of globally available satellite images, such as provided by Sentinel-2, creates new possibilities for researchers to use freely available multi-spectral optical images, with decametric spatial resolution and more frequent revisits for remote sensing applications such as land cover and crop classification (LC\&CC), agricultural monitoring and management, environment monitoring. Existing solutions dedicated to cropland mapping can be categorized based on per-pixel based and object-based. However, it is still challenging when more classes of agricultural crops are considered at massive scale. In this paper, a novel and optimal deep learning model for pixel-based LC\&CC is developed and implemented based on Recurrent Neural Networks (RNN) in combination with Convolutional Neural Networks (CNN) using multi temporal sentinel-2 imagery of central north part of Italy, which has diverse agricultural system dominated by economic crop types. The proposed methodology is capable of automated features extraction by learning time correlation of multiple images, which reduces manual feature engineering and modeling crops phenological stages. Fifteen classes, including major agricultural crops, were considered in this study. We also tested other widely used traditional machine learning algorithms for comparison such as support vector machine SVM, random forest (RF), Kernal SVM, and gradient boosting machine, also called XGBoost. The overall accuracy achieved by our proposed Pixel R-CNN was 96.5\%, which showed considerable improvements in comparison with existing mainstream methods. This study showed that Pixel R-CNN based model offers a highly accurate way to assess and employ time-series data for multi-temporal classification tasks.
\end{abstract}

% keywords can be removed
%\keywords{First keyword \and Second keyword \and More}

\section{Introduction}
Significant increase in population around the globe, demanding increase in  agricultural productivity and thus precise land cover and crop classification and spatial distribution of various crops are becoming significant for governments, policymakers and farmers to improve decision-making process to manage agricultural practices and needs \cite{b1}. Crop maps are produced relatively at large scale, ranging from global\cite{Wu 2014}, countrywide \cite{Jin 2019}, and local level\cite{Yang 2007, Wang 2016}. The growing need for agriculture in the management of sustainable natural resources becomes essential for the development of effective cropland mapping and monitoring \cite{b2}. Group on Earth Observations (GEO), with its Integrated Global Observing Strategy (IGOS), also emphases on an operational system for monitoring global land covers and mapping spatial distribution of crops by using remote sensing imagery. spatial information of the crop maps have been the main source for crop growth monitoring\cite{Huang 2015, Battude 2016, Guan 2016}, water resources managment \cite{Toureiro 2017}, decsion making for policy makers to ensure food security \cite{Yu 2017}.\\
 Satellite and Geographic Information System (GIS) data have been an important source factor in establishing and improving the current systems that are responsible for developing and maintaining land cover and agricultural maps \cite{b3}. Freely available satellite data offers one of the most applied sources for mapping agricultural land and assessing important indices that describe conditions of crop fields \cite{b4}. Recently launched sentinel-2 is equipped with multispectral imager that can provide up to 10m per pixel spatial resolution with the revisit time of 5 days, which offers great opportunity to be exploited in remote sensing domain.\\
Multispectral time series data acquired from MODIS and LANDSAT have been widely used in many agricultural applications such as crop yield prediction \cite{b5}, landcover and crop classification \cite{b6}, leaf area index estimation \cite{b7}, plant height estimation \cite{b8}, vegetation variability assessment \cite{b9} and many more. Two different data sources can also be used together to extract more features that lead to improving results. For example, Landsat-8 and sentinel-1 used together for LC\&CC \cite{b10}.\\
There are some supervised or unsupervised algorithms for mapping cropland using mono or multi-temporal images \cite{b11,b12}. Multi-temporal images have already proven to gain better performance than mono temporal mapping methods \cite{b13,b14}. The imagery used for only key phenological stages s proved to be sufficient for crop area estimation \cite{b18,b15}. It has also found in \cite{b19}, that reducing time-series length affects the average accuracy of the classifier.  Crop patterns were established using Enhanced Vegetation Index derived from 250 meters MODIS-Terra time series data and used to classify some major crops like corn, cotton, and soybean in Brazil \cite{b20}. Centimetric resolution imagery is available at the cost of high price of commercial satellite imagery or with the extensive UAV flight campaigns to cover large area during the whole crop cycle to get better spatial and temporal details. However, most of the studies used moderate spatial resolution(10-30m) freely available satellite imagery for land cover mapping due to their high spectral and temporal resolution which is difficult in case of UAV and high resolution satellite imagery.\\
Other than multispectral time series data, several vegetation indices (VIs) derived from different spectral bands have been exploited and used to enrich the feature space for vegetation assessment and monitoring \cite{b21,b22}. VIs such as normalized difference vegetation index (NDVI), normalized difference water index (NDWI), enhanced vegetation indexes (EVI), textural features, such as grey level co-occurrence matrix (GLCM), statistical features, such as mean, standard deviation, inertial moment are the features more frequently used for crop classification. It is possible to increase the accuracy of the algorithms also using ancillary data such as elevation, census data, road density, or coverage. Nevertheless, all these derived features, along with the phenological metrics involve a huge volume of data, which may increase computational complexity with little improvement in accuracy \cite{b23}. Several feature selection methods have been proposed \cite{b23} to deal with this problem. In \cite{b24}, various features have been derived from MODIS time series and best feature selection has been made using random forest algorithm.\\
 LC\&CC can also be classified as pixel-based or object-based. Object-based image analysis (OBIA), described by Blaschke, that segmentation of satellite images into homogeneous image segments can be achieved with high-resolution sensors \cite{b25}. Various object-based classification has been proposed to produce crop maps using satellite imagery \cite{b26,b27,b28}.  \\
In this work, we proposed a unique deep neural network architecture for LC\&CC, which comprises of Recurrent Neural Network (RNN) that extracts temporal correlations from time series of sentinel-2 data in combination with Convolutional Neural Network (CNN) that analyzes and encapsulate the crops pattern through its filters. The remainder of this paper is organized as follows. Section II briefs about related work done for the LC\&CC along with an overview of RNN and CNN. Section III provides an overview of the raw data collected and exploited during the research. Section IV provides detailed information on the proposed model and the training strategies. Section V contains a complete description of the experiments, results, and discussion along with the comparison with previous state-of-the-art results. Finally, Section VI draws some conclusions. 

%%%%%%%%%%%%%%%%%%%%%%%%%%%%%%%%%%%%%%%%%%
\section{Related work}
\subsection{Temporal feature representation}
There are various studies proposed in the past to address LC\&CC. A more common approach adopted for classification tasks is to extract temporal features and phenological metrics from the VIs time series derived from remotely sensed imagery. There are also some simple statistics and threshold-based procedures used to calculate vegetation related metrics such as Maximum VI and time of peak VI \cite{b29,b30}, which have improved classification accuracy when compared to using only VI as features \cite{b31}.  More complex methods have been adapted to extract temporal features and patterns to address the vegetation phenology \cite{b32}. Further, time series of VI represented by a set of functions \cite{b33}, linear regression \cite{b34}, Markov model \cite{b35} and curve-fitting functions. Sigmoid function has been exploited by \cite{b36,b37}, and achieved better results due to its robustness and ease to derive phenological features for the characterization of vegetation variability \cite{b38}. Although above-mentioned methods of temporal feature extraction offer many alternatives and flexibilities in deployment to assess vegetation dynamics, in practice, there are some important factors such as manually designed model and feature extraction, intra-class variability, uncertain atmospheric conditions, empirical seasonal patterns, which make the selection of such methods more difficult. Thus, an appropriate approach is needed to fully utilize the sequential information from time series of VI to extract temporal patterns.   
As our proposed DNN architecture is based on pixel classification, therefore the following subsections will provide relevant studies and description
\subsection{Pixel based crops classification}
A detailed review of the state-of-the-art supervised
pixel-based methods for land cover mapping was performed in \cite{b39}. It was found that support vector machine (SVM) for mono temporal image classification was the most efficient in terms of overall accuracy (OA) of about 75\%. The second approach was neural networks (NN) based classifier with almost the same OA 74\%.  SVM is complex and resource-consuming for time series multispectral data applications with broad area classification. Another common approach in the remote sensing applications is the random forest (RF)-based classifiers \cite{b40}. Nevertheless, multiple features should be derived to feed the RF classifier for effective use.
Deep Learning (DL) is a branch of machine learning, and it is a powerful tool that is being widely used in solving a wide range of problems related to signal processing, computer vision, image processing, image understanding, and natural language processing \cite{b41}. The main idea is to discover not only the mapping from representation to output but also the representation itself. That is achieved by breaking a complex problem into a series of simple mappings, each described by a different layer of the model, and then composing them in a hierarchical fashion. A large number of state of the art models, frameworks, architecture, and benchmark databases of reference imagery exist for image classification domain. 
\subsection{Recurrent neural network (RNN)}
Sequence data analysis is an important aspect in many domains, ranging from natural language processing, handwriting recognition, image captioning, to robot automation. In recent years, Recurrent Neural Networks (RNNs) have proven to be a fundamental tool for sequence learning \cite{b48}, allowing to represent information from context window of hundreds of elements. Moreover, the research community has, over the years, come up with different techniques to overcome the difficulty of training over many time steps. For example, long short-term memory (LSTM) \cite{b49} and gated recurrent unit (GRU) \cite{b51}, based architectures have proven ground-breaking achievements \cite{b52,b53}, in comparison to standard RNN models. In remote sensing applications, RNNs are commonly used when sequential data analysis is needed. For example, Lyu et al. \cite{b58} employed RNN to use sequential properties such as spectral correlation and intra-bands variability of multispectral data. They further used LSTM model to learn a combined spectral-temporal feature representation from an image pair acquired at two different dates for change detection \cite{b59}
\subsection{Convolutional neural network (CNN)}

Convolutional Neural Networks (CNNs) date back decades \cite{b60}, emerging from the study of the brain's visual cortex \cite{b61} and classical concepts of computer vision theory \cite{b62}, \cite{b63}. Since the 1990s, these have been applied successfully in image classification \cite{b60}. However, due to technical constraints such as mainly lack of hardware performance, the large amount of data, and theoretical limitations, CNNs did not scale to large applications. Nevertheless, Geoffrey Hinton and his team demonstrated at the annual ImageNet ILSVRC \cite{b64} competition the feasibility to train large architectures capable of learning several layers of features with increasingly abstract internal representations \cite{b65}. Since that breakthrough achievement, CNNs became the ultimate symbol of the Deep Learning \cite{b41} revolution, incarnating all those concepts that underpin the entire novel movement.\\  In recent years, DL was widely used in data mining and remote sensing applications. In particular, image classification studies exploited several DL architectures due to their flexibility in feature representation, and automation capability for end-to-end learning. In DL models, features can be automatically extracted for classification tasks without feature crafting algorithms by integrating autoencoders \cite{b83,b84}. 2D CNNs have been broadly used in remote sensing studies to extract spatial features from high-resolution images for object detection and image segmentation \cite{b85,b86,b87}. In crop classification, 2D convolution in the spatial domain performed better than 1D convolution in the spectral-domain \cite{b88}. These studies formed multiple convolutional layers to extract spatial and spectral features from remotely sensed imagery. 

%%%%%%%%%%%%%%%%%%%%%%%%%%%%%%%%%%%%%%%%%%
\section{Study area and data}
\begin{figure*}[htpb]
\centering
 \includegraphics[scale=0.35]{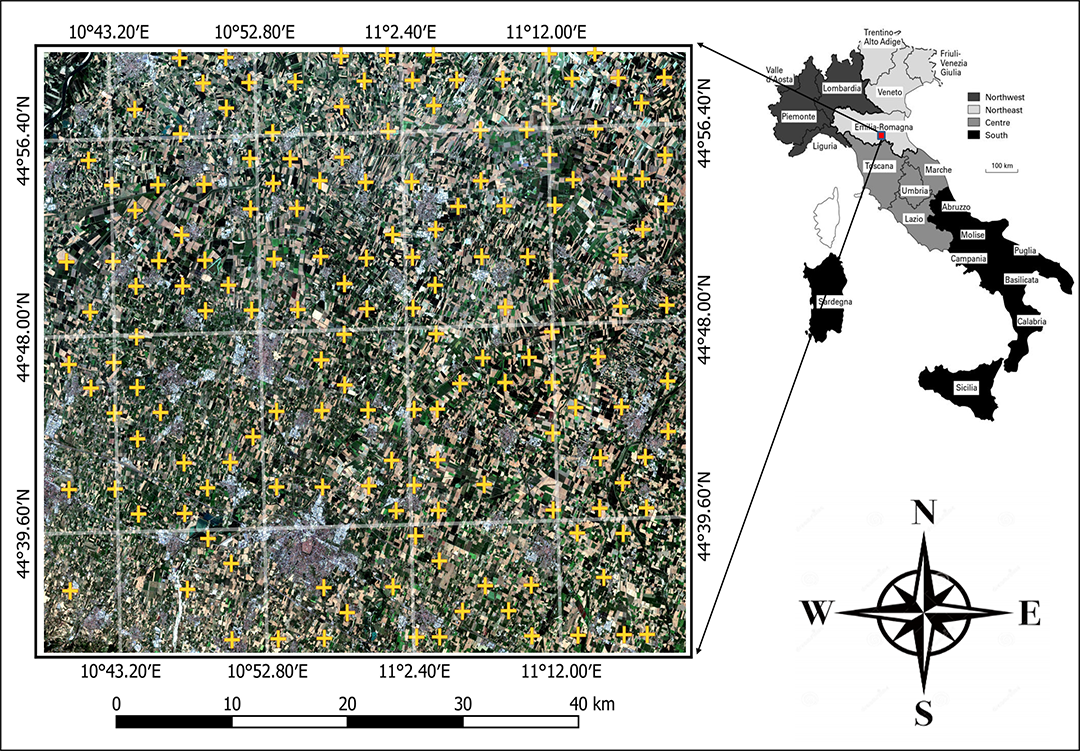}
  \caption {The study site is located in Carpi, region Emilia-Romagna is shown with the geo-coordinates (WGS84). RGB image composite derived from sentinel-2 imagery  acquired in August-2015 is shown and the yellow marker showing geo-locations of ground truth land cover extracted from Land Use and Coverage Area frame Survey (LUCAS-2015).}
\label{fig:fig1}
\end{figure*}

The study site near Carpi, Emilia-Romagna, situated in center-north part of Italy with central coordinates$44^{\circ} 47'01''N, 10^{\circ} 59'37''E$ was considered for LC\& CC shown in Figure \ref{fig:fig1}. The Emilia-Romagna region is one of the most fertile plains of Italy. An area almost 2640 $km^2$ was considered, which covers diverse crop land. The major crop fields in this region are Maize, Lucerne, Barley, Wheat, and Vineyards. The yearly averaged temperature and precipitation are $14^{\circ} C$ and 843 $mm$ for this region. Most of the farmers practice single cropping in this area. 

% Table generated by Excel2LaTeX from sheet 'Sheet1'
\begin{table}[htp]
  \centering
  \caption{Bands used in this study.}
    \begin{tabular}{p{10.215em}p{20.645em}p{5em}c}
    \toprule
    Bands used & Description & Central wavelength ($\mu$m) & \multicolumn{1}{p{4.215em}}{Resolution (m)} \\
    \toprule
    Band 2 & Blue & \multicolumn{1}{c}{0.49} & 10 \\
    \hline
    Band 3 & Green & \multicolumn{1}{c}{0.56} & 10 \\
    \hline
    Band 4 & Red  & \multicolumn{1}{c}{0.665} & 10 \\
    \hline
    Band 8 & Near infrared & \multicolumn{1}{c}{0.705} & 10 \\
    \hline
    NDVI & (Band8-Band4)/ (Band8+Band4) &     -    & 10 \\
    \bottomrule
    \end{tabular}%
  \label{tab:tab1}%
\end{table}%

To know about the spatial distribution of crops, we deeply studied Land Use Cover Area frame statistical Survey (LUCAS) and extracted all the information we need for ground truth data. LUCAS was carried out by Eurostat to be able to monitor the agriculture, climate change, biodiversity, forest, and water for almost all over the Europe \cite{b89}.

 The technical reference document of LUCAS-2015 was used to prepare the ground truth data. Microdata that contains spatial information of crops and several land cover types along with the geo-coordinates for the considered region was imported in Quantum Geographic information system (QGIS) software, an Open-source software used for visualization, editing, analysis of geographical data. The selection of pixel was made manually by overlapping images and LUCAS data, so a proper amount of ground truth pixels were extracted for training and testing the algorithm. The sentinel-2 mission consists of twin polar-orbiting satellites launched by European Space Agency (ESA) in 2015 and can be used in various application areas such as land cover change detection, natural disaster monitoring, forest monitoring, and most importantly in agricultural monitoring and management
\cite{b91}. 
% Table generated by Excel2LaTeX from sheet 'Sheet1'
\begin{table}[htbp]
  \centering
  \caption{Sentinel-2 data acquisition.} 
    \begin{tabular}{ccp{6.215em}c}
    \toprule
    \multicolumn{1}{p{4.215em}}{Date} & \multicolumn{1}{p{4.215em}}{Doy} & Sensing Orbit \# & \multicolumn{1}{p{4.93em}}{Cloud pixel percentage} \\
    \toprule
    7/4/2015 & 185  & 22-Descending & 0 \\
    \hline
    8/3/2015 & 215  & 22-Descending & 0.384 \\
    \hline
    9/2/2015 & 245  & 22-Descending & 4.795 \\
    \hline
    9/12/2015 & 255  & 22-Descending & 7.397 \\
    \hline
    10/22/2015 & 295  & 22-Descending & 7.606 \\
    \hline
    2/19/2016 & 50   & 22-Descending & 5.8 \\
    \hline
    3/20/2016 & 80   & 22-Descending & 19.866 \\
    \hline
    4/29/2016 & 120  & 22-Descending & 18.61 \\
    \hline
    6/18/2016 & 170  & 22-Descending & 15.52 \\
    \hline
    7/18/2016 & 200  & 22-Descending & 0 \\
    \bottomrule
    \end{tabular}%
  \label{tab:tab2}%
\end{table}%

\begin{figure*}[htp]
\centering
 \includegraphics[scale=0.35]{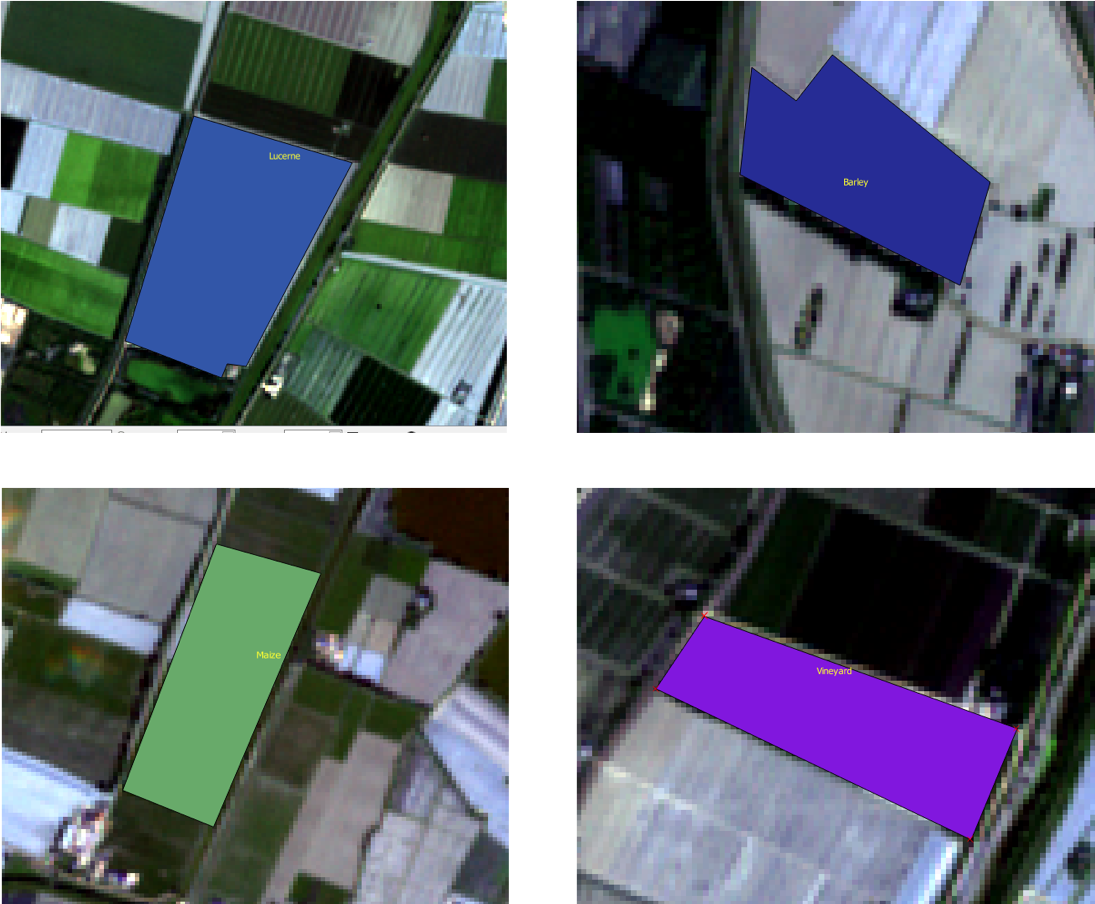}
  \caption {Few examples of zoomed in part of crop classes considered as ground truth. Shape files are used to extract pixels for reference data. }
\label{fig:}
\end{figure*}
It is equipped with multi-spectral optical sensors that capture 13 bands of different wavelengths. We used only high-resolution bands that have 10 meter/pixel resolution shown in Table \ref{tab:tab1}. It also has high revisit time ( ten days at the equator and five days with twin satellites (Sentinel-2A, Sentinel-2B). It became more popular in remote sensing community due to fact that it possesses various key features such as, free access to data products available at ESA Sentinel Scientific Data Hub with reasonable spatial resolution (which is 10m for Red, Green, Blue and Near Infrared bands), high revisit time and reasonable spectral resolution among other available free data sources. In our study, we used ten multitemporal sentinel-2 images reported in Table. \ref{tab:tab2}, which are well co-registered from July-2015 to July-2016 with close to zero cloud coverage. The initial image selection was performed based on the cloudy pixel contribution at the granule level. This pre-screening was followed by further visual inspection of scenes and resulted in a multi-temporal layer stack of ten images. Sentinel Application Platform (SNAP) v5.0 along with sen2core v 2.5.1 were used to apply radiometric and geometric corrections to acquire Bottom of Atmosphere (BOA) Level 2A images from Top of Atmosphere (TOA) Level 1C. Further details about geometric, radiometric correction algorithms used in sen2cor can be found in \cite{b92}. Bands with 10 meter/pixel along with the derived Normalized Difference Vegetation Index (NDVI) were used for experiments, as shown in Table \ref{tab:tab1}.

%%%%%%%%%%%%%%%%%%%%%%%%%%%%%%%%%%%%%%%%%%
\section{Convolutional and recurrent neural networks for pixel-based crops classification}
\subsection{Formulation}
A single multi-temporal, multi-spectral pixel can be represented as a two-dimensional matrix $X^{(i)}\in\mathbb{R}^{t*b}$ where $t$ and $b$ are the number of time steps and spectral bands, respectively. Our goal is to compute from $X^{(i)}$ a probability distribution $F(X^{(i)})$ consisting of $K$ probabilities, where $K$ is equal to the number of classes. In order to achieve this objective, we propose a compact representation learning architecture composed of three main building blocks: 
\begin{itemize}
    \item \textbf{Time correlation representations} - this operation extracts temporal correlations from multi-spectral, temporal pixels $X^{(i)}$ exploiting a sequence-to-sequence recurrent neural network based on Long Short-Term Memory (LSTM) cells. A final Time-Distributed layer is used to compress and maintain a sequence like structure, preserving the multidimensionality nature of the data. In this way, it is possible to take advantage of temporal and spectral correlations simultaneously.
\begin{figure}[h]
\centering
 \includegraphics[scale=0.35]{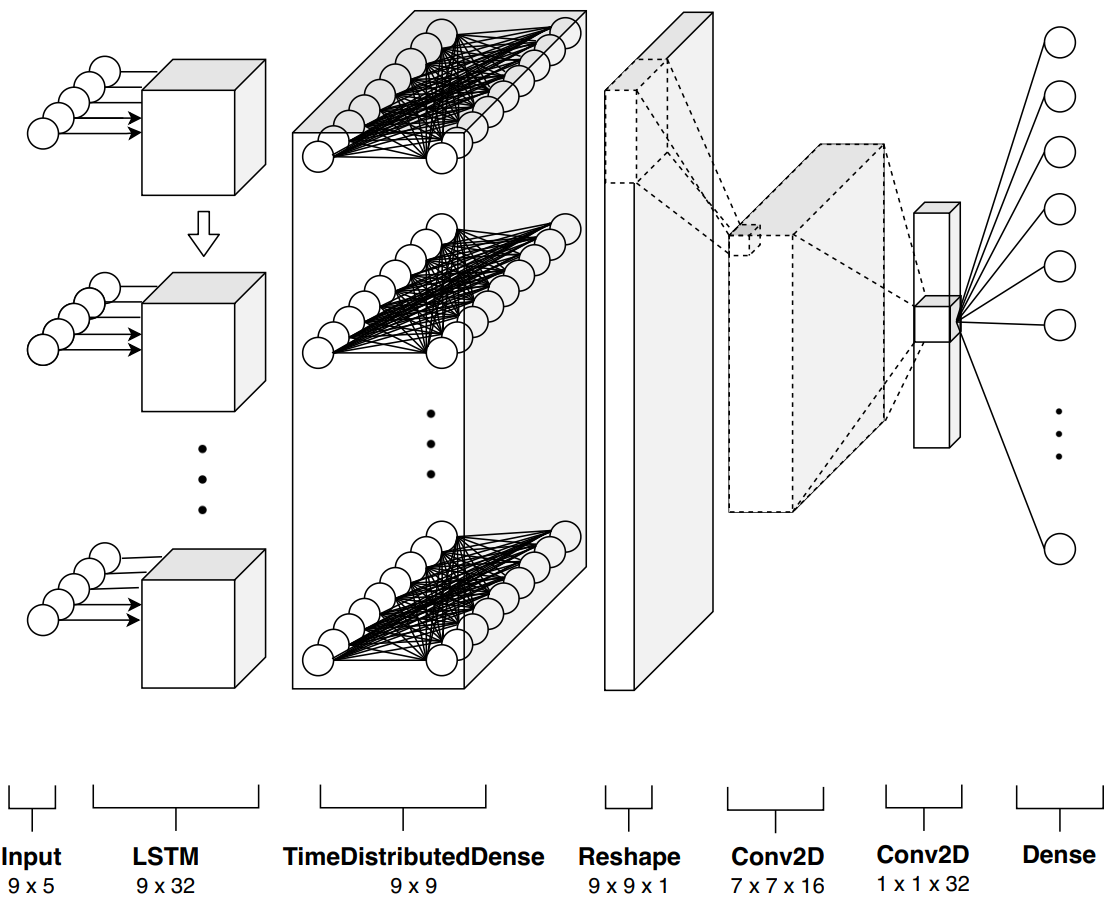}
  \caption {An overview of the Pixel R-CNN model used for classification. Given a multi-temporal, multi-spectral input pixel $X^{(i)}$, the first layer of LSTM units extracts sequences of temporal patterns. A stack of convolutional layers hierarchically processes the temporal information.}
\label{fig:fig2}
\end{figure}
    \item \textbf{Temporal pattern extraction} - this operation consists of a series of convolutional operations followed by rectifier activation functions that non linearly maps each elaborated temporal and spectral patterns onto high dimensional representations. So, RNN output temporal sequences are processed by a subsequent cascade of filters, which in a hierarchical fashion, extracts essential features for the successive stage.
    \item \textbf{Multiclass classification} - this final operation maps the feature space with a probability distribution $F(X^{(i)})$ with $K$ different probabilities, where $K$, as previously stated, is equal to the number of classes.
\end{itemize}

The comprehensive pipeline of operations constitutes a lightweight, compact architecture able to non-linearly map multi-temporal information with its intrinsic nature, achieving results considerably better than previous state-of-the-art solutions. Human brain mental imagery studies \cite{b93}, where images are a form of internal neural representation, inspired the presented architecture. Moreover, the joint effort of RNN and CNN distributes the knowledge representation through the entire model, exploiting one of the most powerful characteristics of deep learning known as distributed learning.  An overview of the overall model, dubbed Pixel R-CNN, is depicted in Fig. \ref{fig:fig2}. Each pixel is extracted contemporary from all images taken at different time steps $t$ with all its spectral bands $b$. In this way, it is possible to create an instance $X^{(i)}$, which can feed the first layer of the network. Firstly, the model extracts temporal representations from the input sample. Subsequently, these temporal features are further enriched by the convolutional layers that patterns in a hierarchical manner. The overall model act as a function $F(X^{(i)})$ that map the input sample with its related probabilities $K$. So, evaluating the probability distribution is possible to identify the class of belonging to the input sample.

It is worth to notice that this model is known as unrolled through time representation. Indeed, only after all time steps have been processed, the CNN is able to analyze and transform the temporal pattern. In the following subsections, we are going to describe in detail each individual block.
\subsubsection{Time correlation representation}
\begin{figure}[!b]
\centering
 \includegraphics[scale=0.28]{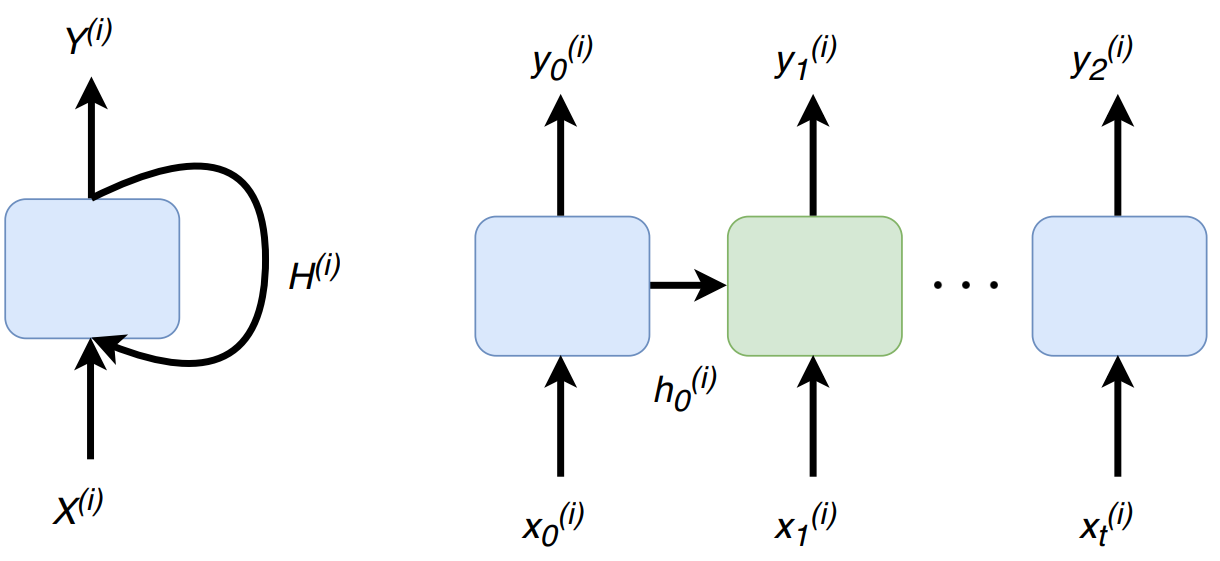}
  \caption {A recurrent layer and its unrolled through time representation. A multi-temporal, multi-spectral pixel $X^{(i)}$ is made by a sequence of time steps, $x^{(i)}_t$, that along the previous output $h^{(i)}$ feed the next iteration of the network.}
\label{fig:fig3}
\end{figure}

\begin{figure}[t!]
\centering
 \includegraphics[scale=0.5]{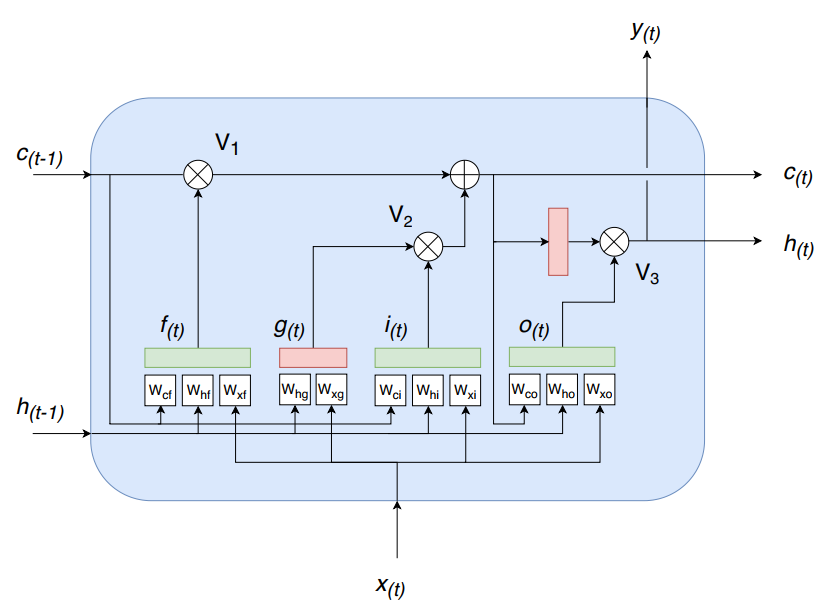}
  \caption {LSTM with peephole connections. A time step $t$ of a multi-spectral pixel $x^{(i)}_t$ is processed by the memory cell which decides what to add and forgot in the long-term state $c_{(t)}$ and what discard for the present state $y_t^{(i)}$.}
\label{fig:fig4}
\end{figure}

Nowadays, a popular strategy in time series data analysis is the use of RNNs that have proven excellent results in many fields of the application over the years. Looking at the simplest possible RNN shown in Fig. \ref{fig:fig3}, composed of just one layer, it looks very similar to a feedforward neural network, except it also has a connection going backward. Indeed, the layer is not only fed by an input vector $x^{(i)}$, but it also receives $h^{(i)}$ (cell state), which is equal to the output neuron itself, $y^{(i)}$. So, at each time step $t$, this recurrent layer receives an input 1-D array $x^{(i)}_t$ as well as its own output from the previous time step, $y_{(t-1)}^{(i)}$. In general, since the output of a recurrent neuron at time step $t$ is a function of all inputs from previous time steps, it has, intuitively, a sort of memory that influences all successive outputs. In this example, it is straightforward to compute a cell's output, as shown in Eq. \ref{eq1}.
	
\begin{equation}\label{eq1}
     y_t^{(i)} = \phi(x^{(i)}_t\cdot W_x+ y^{(i)}_{(t-1)}\cdot W_y+b)
\end{equation}

where, in the context of this research,  $x_t^{(i)} \in\mathbb{R} ^{(1*b)}$  is a single time step of a pixel with $n_{inputs}$ equal to the number of spectral bands b. $y_t^{(i)}$ and $y_{(t-1)}^{(i)}$  are the output of the layer at time $t$ and ${t-1}$, respectively, $W_x$ and $W_y$ are the weights matrices. It is important to point out that $y_t$ as $x_t^{(i)}$ are vectors and they can have an arbitrary number of elements, but the representation Fig. \ref{fig:fig3} does not change. Simply, all neurons are hidden in the depth dimension.
 Unfortunately, the basic cell just described suffer from major limitations, but most of all are the fact that, during training, the gradient of the loss function gradually fades away. For this reason, for the time correlation representation, we adopted a more elaborated cell known as peephole LSTM unit, see Fig. \ref{fig:fig4}.  That is an improved variation of the concept proposed in 1997 by Sepp Hochreiter and Jurgen Schmidhuber \cite{b49}. The key idea is that the network can learn what to store in a long-term state, $c_{(t)}$ what to throw away and what to use for the current state $h_{(t)}$ and $y_{(t)}$ that, as for the basic unit, are equal. That is performed with simple element-wise multiplications working as $"valves"$ for the fluxes of information. Those elements, $V_1$, $V_2$ and $V_3$ are controlled by fully connected (FC) layers that have as input the current input state $x_{(t)}$ and the previous short-term memory term $h_{(t-1)}$. Moreover, for the peephole LSTM cell, the previous long-term state $c_{(t-1)}$ is added as an input to the FC of the forgot gate,$V_1$, and the input gate, $V_2$. Finally, the current long-term state $c_t$ is added as an input to the FC of the output gate. All "gates controllers" have sigmoid as activation functions (green boxes) instead of tanh ones to process the signals themselves (red boxes). So, to summarize, a peephole LSTM block has three signals as input and output; two are the standard input state $x_{(t)}$ and cell output $y_{(t)}$. Instead, $c$ and $h$ are the long and short-term state, respectively, that the unit, utilizing its internal controllers and valves, can feed with useful information. Formally, as for the basic cell seen before, Eq (2). Eq (7). summarizes how to compute the cell's long-term state, its short-term state, and its output at each time step for a single instance.

\begin{equation}\label{eq2}
i_{(t)} = \sigma ( W^T_{ci}\cdot c_{(t-1)} + W^T_{hi}\cdot h_{(t-1)} + W^T_{xi}\cdot  x^{(i)}_{(t)} + b_i )
\end{equation}
\begin{equation}\label{eq3}
     f_{(t)} = \sigma( W^T_{cf}\cdot c_{(t-1)} + W^T_{hf}\cdot h_{(t-1)} + W^T_{xf} \cdot x^{(i)}_{(t)} + b_f )
\end{equation}
\begin{equation}\label{eq4}
     o_{(t)} = \sigma( W^T_{co}\cdot c_{(t)} + W^T_{ho}\cdot h_{(t-1)} + W^T_{xo}\cdot  x^{(i)}_{(t)} + b_o )
\end{equation}
\begin{equation}\label{eq5}
     g_{(t)} = \tanh( W^T_{hg}\cdot h_{(t-1)} + W^T_{xg}\cdot  x^{(i)}_{(t)} + b_g  )
\end{equation}
\begin{equation}\label{eq6}
     c_{(t)} = f_{(t)}\otimes c_{(t-1)} +  i_{(t)}\otimes g_{(t)} 
\end{equation}
\begin{equation}\label{eq7}
     y_{(t)} = h_{(t)} =o_{(t)}\otimes \tanh(c_{(t)} )    
\end{equation}

\begin{figure}[t!]
\centering
 \includegraphics[scale=0.8]{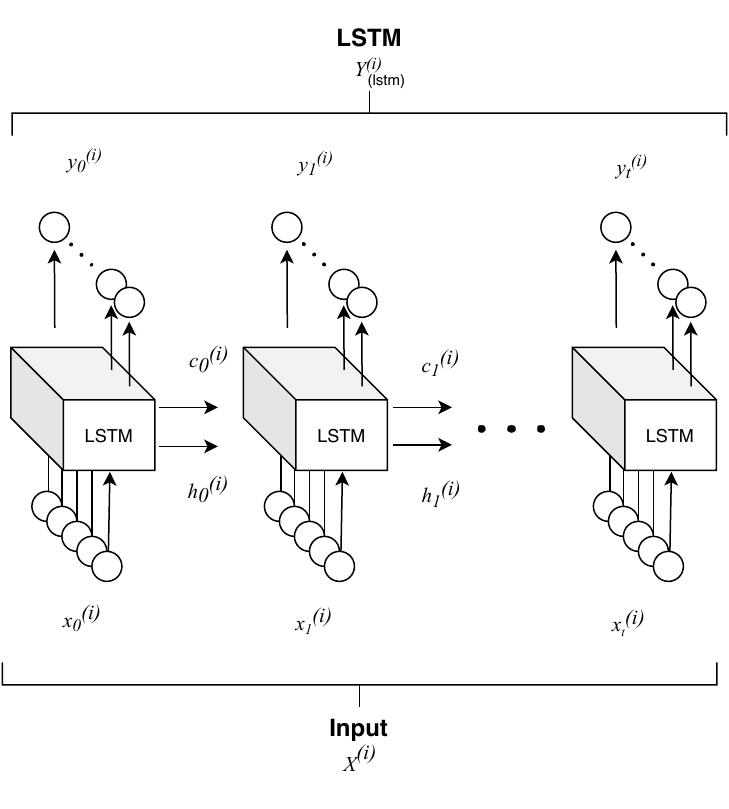}
  \caption {Pixel R-CNN first layer for time correlations extraction. Peephole LSTM cells extracts temporal representations from input instances $X^{(i)}\in\mathbb{R}^{t*b}$. The output matrix $Y^{(i)}_{(lstm)}$ feeds a TimeDistributedDense layer, that preserves the multidimensional nature of the processed data extracting multi-spectral patterns.}
\label{fig:figx}
\end{figure}

In conclusion, multi-temporal, multi-spectral pixel $X^{(i)}$ is processed by a first layer of LSTM peephole cells obtaining a cumulative output $Y_{(lstm)}^{(i)}$. Finally, a TimeDistributedDense layer is applied which executes simply a Dense function across every output over time, using the same set of weights, preserving the multidimensional nature of the processed data Eq.(8). In Fig. \ref{fig:figx} is presented a graphical representation of the first layer of the network. LSTM cells extract temporal representations from input samples $X^{(i)}$ with  $x_t^{(i)} \in\mathbb{R} ^{(1*b)}$ as columns. The output matrix $Y_{(lstm)}^{(i)}$ feeds the subsequent TimeDistributedDense layer.
\begin{equation}\label{eq8}
    F_{timeD}(F_{lstm}(X^{(i)}))= ( W \cdot Y_{(lstm)}^{(i)} + B )  
\end{equation}
\subsubsection{Temporal patterns extraction}
\begin{figure}[t!]
\centering
 \includegraphics[scale=0.8]{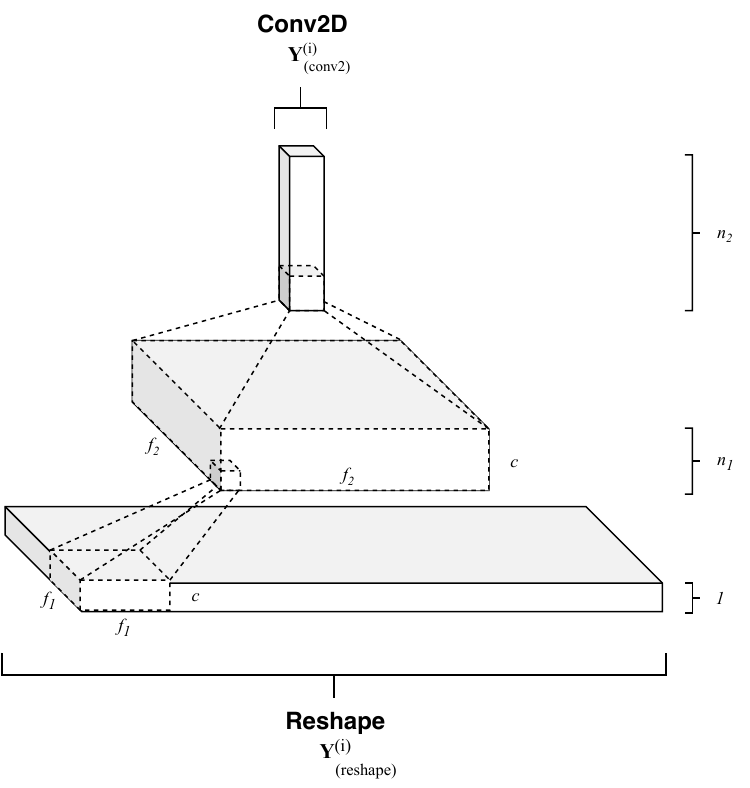}
  \caption {Pixel R-CNN convolutional layers. Firstly, output from TimeDistributedDense layer $Y_{(timeD)}^{(i)}$ is reshaped in a 3-dimensional tensor $\textbf{Y}^{(i)}_{(reshape)}$ and then it feeds a stack of two convolutional layers that progressively reduce the first two dimensions, gradually extracting higher level represenetations.}
\label{fig:figxii}
\end{figure}

The first set of layers extract a 2-dimensional tensor $Y_{(timeD)}^{(i)}$ for each instance. In the second operation, after a simple reshaping operation in order to increase the dimensionality of the input tensor from two to three and beeing able to apply the following operations, we map each of these 3-dimensional array $\textbf{Y}^{(i)}_{(reshape)}$ into an higher dimensional space. That is accomplished by two convolutional operations, built on top of each other, that hierarchically apply learned filters, extracting gradually more abstract representations. More formally, the temporal patterns extraction is expressed, for example, for the first convolutional layer, as an operation $F_{conv1}$
\begin{equation}\label{eq9}
    F_{conv1} (F_{timeD} (F_{lstm}(X^{(i)})))= \max( 0, W_1 \ast \textbf{Y}_{(reshape)}^{(i)} + B_1 )  
\end{equation}
where $W_1$ and $B_1$ represent filters and biases, respectively, and '$\ast$' is the convolutional operation. $W_{1}$, contains $n_{1}$ filters with kernel dimension $f_1$ x $f_1$ x $c$, where $f_1$ is the spatial size of a filter and $c$ is the number of input channels. As common for CNN, rectified linear unit (ReLU), max(0,x), has been chosen as activation function for both layers units. In Fig. \ref{fig:figxii} is depicted a graphical scheme of this section of the model. So, summarizing, output matrix $Y_{(timeD)}^{(i)}$ of the TimeDistributedDense layer, after adding an extra dimension, feeds a stack of two convolutional networks that progressively reduce the first two dimensions, gradually extracting higher level representations and generating high dimensional arrays. Moreover, being the $n_1$ and $n_2$ filters shared across all units, the same operation carried out with a similarly-sized dense fully connected layers would require a much greater number of parameters and computational power. Instead, the synergy of RNN and CNN opens the possibility to elaborate the overall temporal canvas in an optimize and efficient way.
\subsubsection{Multiclass classification}
In the last stage of the network, the extracted feature tensor $\textbf{Y}_{(conv2)}^{(i)}$, after removing the extra dimensions with a simple flatten operation, is mapped to a probability distribution consisting of $K$ probabilities, where $K$ is equal to the number of classes.
This is achieved by a weighted sum followed by a softmax activation function:
\begin{equation}\label{eq10}
     \hat{p}_{k} = \sigma(s(x))_k = \frac{\exp{s_k(x)}} {\sum_{j=1}^{K} \exp{s_j(x)} } for\ j=1..,K
\end{equation}
where $s(x)$=  $W^T$.$y_{(flatten-conv2)}^{(i)}$ + $B$ is a vector containing the scores of each class for the input vector $y_{(flatten-conv2)}^{(i)}$, that after the flatten operaton is a 1-dimensional array. Weights W and bias B are learned, during the training process, in such a way to classify arrays of the high dimensional space into the K different classes. So, $\hat{p}_k$ is the estimated probability that the extracted feature vector $y_{(flatten-conv2)}^{(i)}$ belongs to class k given the scores of each class for that instance.
\subsection{ Training}
Learning the overall mapping function F requires the estimation of all network parameters $\Theta$ of the three different model parts. This is simply achieved through minimizing the loss between each pixel class prediction $F(X ^{(i)})$  and the corresponding ground truth $y^{(i)}$ with a supervised learning strategy. So, given a data set with $n$ pixel samples $\left\{ X_i \right\}$and the respective true classes set$\left\{ y _i \right\}$, we use categorical cross-entropy as the loss function:
\begin{equation}\label{eq11}
    J( \Theta)= -1/n \sum_{i=1}^{n} \sum_{k=1}^{K}y_k^{(i)} \log( \hat{p}_k^{(i)}) 
\end{equation}
where $y_k^{(i) }$ cancels all classes loss except for the true one. Equation (11). is minimized using AMSGrad optimizer \cite{b94}, an adaptive learning rate method which modifies the basic ADAM optimizer \cite{b95} algorithm. The overall algorithm update rule without the debiasing step is:

\begin{equation}\label{eq12}
    m_t = \beta_1 m_{t-1} + ( 1-\beta_1)g_t 
\end{equation}
\begin{equation}\label{eq13}
  v_t = \beta_2 v_{t-1} +  ( 1-\beta_2)g_t^2  
\end{equation}
\begin{equation} \label{eq14}
  \hat{v}_t = \max ( \hat{v}_{t-1},v_t)
\end{equation}
\begin{equation}\label{eq15}
  \theta_{t+1} = \theta_t - \frac{\eta}{\sqrt{\hat{v}_t+\epsilon}} m_t
\end{equation}
 Equation (12). and Eq. (13). are the exponential decay of the gradient and gradient squared, respectively. Instead, with the Eq. (14)., keeping a higher $v_t$ term results in a much lower learning rate, $\eta$ , fixing the exponential moving average and preventing to converge to a sub-optimal point of the cost function. Moreover, we use a technique known as cosine aneling in order to cyclically vary the learning rate value between certain boundary values \cite{b96}. This value can be obtained with a preliminary training procedure, linearly increasing the learning rate while observing the value of the loss function. Finally, we employ, as only regularization methodology, "Dropout" \cite{b97} in the time representation stage, inserted between the LSTM and Time-Distributed layer. This simple tweak allows training a more robust and resilient to noise temporal patterns extraction stage. Indeed, forcing CNN to work without relying on certain temporal activations can greatly improve the abstraction of the generated representations distributing the knowledge across all available units. 

%%%%%%%%%%%%%%%%%%%%%%%%%%%%%%%%%%%%%%%%%%
\section{Experimental Results and discussion}
We first processed raw data in order to create a set of \textbf{n} pixel samples   $\mathbb{X}$ =$ \{\textbf{X}_i\}$ with the related ground truth labels $\mathbb{Y}$=$\{y_i\}$.
Then, in order to have a visual inspection of the data set, principal component analysis (PCA), one of the most popular dimensionality reduction algorithms, have been applied to project the training set onto a lower tri-dimensional hyperplane. Finally, quantitative and qualitative results are discussed with a detail description of the architecture settings.

\subsection{Training data}
Sample pixels require to be extracted from the raw data and then reordered in order to feed the devised architecture. Indeed, the first RNN stage requires data points to be collected in slices of time series.
\begin{figure}[h]
\centering
 \includegraphics[scale=0.80]{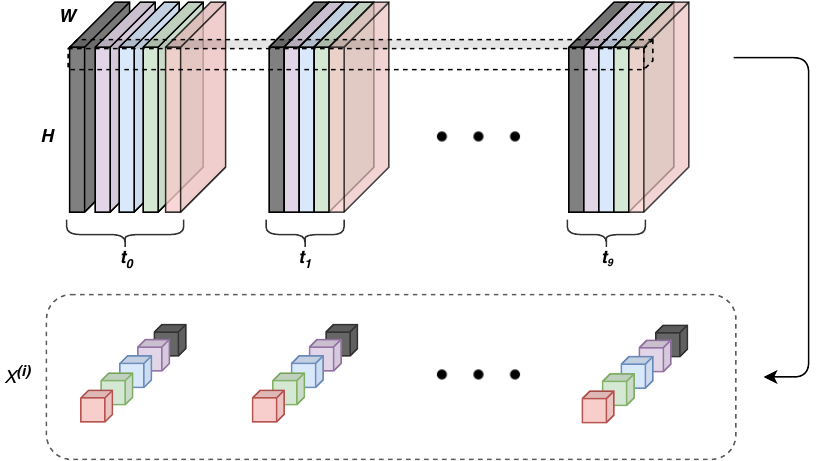}
  \caption {Overview of the tensor $\textbf{X} \in \mathbb{R}^{i\times t\times b}$ generation. The first dimension $i$ represents the collected instances $X^{(i)}$, the second $t$ the different time steps, and finally the last one $b$ the five spectral bands, red, green, blue, near-infrared and NDVI. On the top, labeled pixels are extracted simultaneously, from all-time steps and bands starting from the raw satellite images. Then, $X_{i,j,k}$ are reshaped in order to set up the $X^{(i)}=\textbf{X}_{i,:,:}$ of the data set tensor \textbf{X}.}
\label{fig:fig5}
\end{figure}
So, we separated labeled pixels from raw data, and we divided them in chunks of data, forming a tri-dimensional tensor $\textbf{X} \in \mathbb{R}^{i\times t\times b}$ for the successive pre-processing pipeline. In Fig. \ref{fig:fig5}, a visual representation of the data set tensor \textbf{X} generation, where fixing the first dimension \textbf{X}$_{i,:,:}$ there are the individual pixel samples  $X^{(i)}$ with $t=9$ time steps and $b=5$ spectral bands. It is worth to notice that the number of time steps and bands are completely an arbitrary choice dictated by the raw data availability.

% Table generated by Excel2LaTeX from sheet 'Sheet1'
\begin{table}[htbp]
  \centering
  \caption{Land cover types contribution in the reference data.}
    \begin{tabular}{p{6.07em}cc}
    \toprule
    \textbf{Class} & \multicolumn{1}{p{4.215em}}{\textbf{Pixels}} & \multicolumn{1}{p{4.215em}}{\textbf{Percentage}} \\
    \midrule
    Tomatoes & 3020  & 3.20\% \\
    Artificials & 9343  & 10.14\% \\
    Trees & 7384 & 8.01\% \\
    Rye   & 4382  & 4.75\% \\
    Wheat & 12826 & 13.92\% \\
    Soya  & 5836  & 6.33\% \\
    Apple & 849   & 0.92\% \\
    Peer  & 495   & 0.53\% \\
    Temp Grass & 1744  & 1.89\% \\
    Water & 2451  & 2.66\% \\
    Lucerne & 17942 & 19.47\% \\
    Drum Wheat & 1188  & 1.28\% \\
    Vineyard & 6110  & 6.63\% \\
    Barley & 2549  & 2.76\% \\
    Maize & 15997 & 17.37\% \\
    \textbf{Total} & 92116 & 100\% \\
    \bottomrule
    \end{tabular}%
  \label{tab:tab3}%
\end{table}%

Subsequently, we adopted a simple pipeline of two steps to pre-process the data. Stratified sampling has been applied in order to divide the data set tensor \textbf{X}, with shape (92116, 9, 5), in a training and test set. Due to the natural unbalanced number of instances per class present in the data set Table. \ref{tab:tab3}, this is an important step in order to preserve the same percentage in the two sets. After selecting a split percentage for the training of 60\%, we obtained two tensors \textbf{X}$_{train}$ and \textbf{X}$_{test}$ with shape (55270, 9, 5) and (36846, 9, 5), respectively. Secondly, as common practice, in order to facilitate the training, we adopted standard scaling, $(x-\mu)/\sigma $, to normalize the two sets of data points.
\begin{figure}[h]
\centering
 \includegraphics[scale=1]{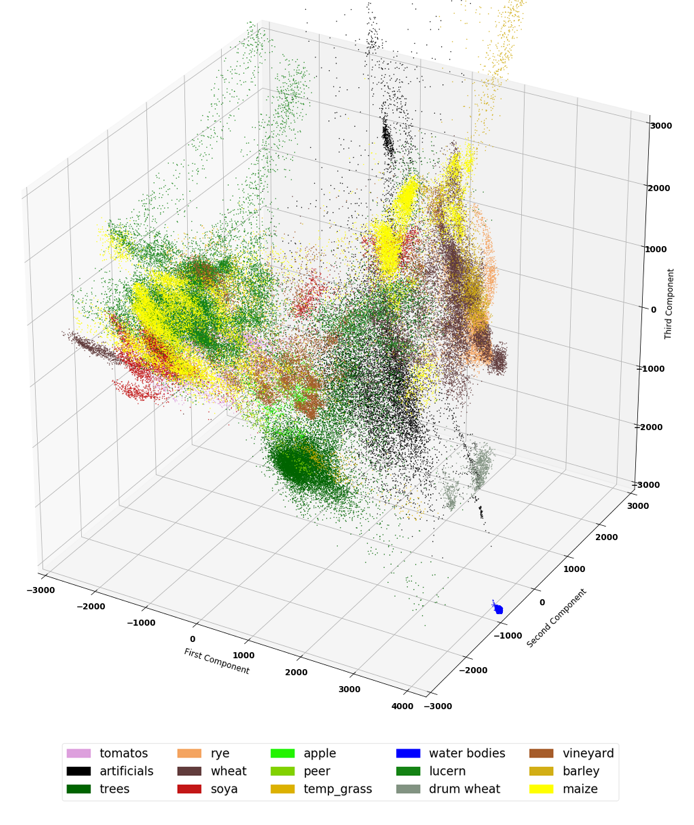}
  \caption {Visual representation of the data points projected in the tri-dimensional space using PCA. The three principal components took into account preserve 64.5\% of the original data set variance.}
\label{fig:fig6}
\end{figure}
\subsection{Dataset Visualization}
\begin{figure}[h]
\centering
 \includegraphics[scale=0.3]{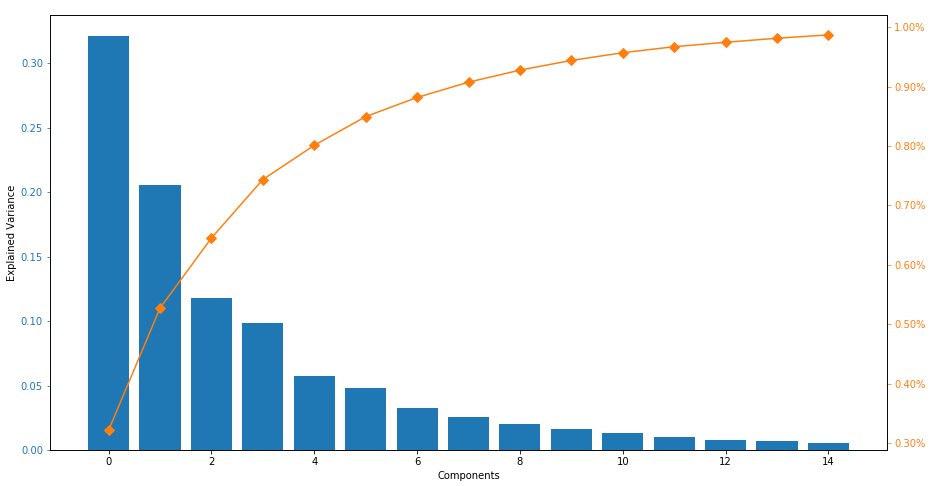}
  \caption {Pareto Chart of the explained variance as a function of the number of components.}
\label{fig:fig7}
\end{figure}

In order to explore and visualize the generated set of points, we exploit Principle Component Analysis (PCA), reducing the high dimensionality of the data set. For this operation, we considered the components t and b as features of our data points. So, applying Singular Value Decomposition (SVD) and then selecting the first three principal components, $W_d=(c_1,c_2,c_3)$, it was possible to plot the different classes in a tri-dimensional space, having a visual representation of the projected data points.
In Fig. \ref{fig:fig6} the projected data points are plotted in tri-dimensional space. Except for water bodies, it is worth to point out how much intra-class variance is present. Indeed, most of the classes lay on more than one hyperplane, demonstrating the difficulty of the task and the studied data set. Finally, it was possible to analyze the explained variance ratio varying the number of dimensions. From Fig. \ref{fig:fig7}, it is worth to notice that approaching higher components, the explained variance trend stops growing fast. So, that can be considered as the intrinsic dimensionality of the data set. Due to this fact, it is reasonable to assume that reducing the number of time steps would not significantly affect the overall results.

\subsection{Experimental settings}
In this section, we examine the settings of the final network architecture. The basic Pixel R-CNN model, shown in Fig. \ref{fig:fig2}, is the result of a careful design aimed at obtaining the best performance in terms of accuracy and computational cost. Indeed, the final model is a lightweight model with 30,936 trainable parameters (less than 1 MB), fast and more accurate than the existing state-of-the-art solutions. With the suggested approach, we employed only an RNN layer with 32 output units for each peephole LSTM cell randomly turned off, with a probability $p$ = 0.2, by a Dropout regularization operation. For all experiments, peephole LSTM has shown an improvement in overall accuracy around 0.8\% over standard LSTM cells. Then, Time Distributed Dense transforms $Y_{(lstm)}^{(i)}$ in a 9x9 square matrix that feed a stack of two CNN layers with a number of features $n_1$ = 16 and $n_2$ = 32, respectively. The first layer as a filter size of $f_1$=3  and the second one $f_2$ = 7  producing a one-dimensional output array. Finally, a fully connected layer with SoftMax activation function maps $Y_{(conv2)}^{(i)}$ to the probability of the $K=15$ different classes. Except for the final layer, we adopted ReLU as activation functions.\ 
In order to find the best training hyperparameters for the optimizer, we used 10\% of the training set to perform a random search evaluation, with few epochs, in order to select the most promising parameters. Then, after this first preliminary phase, the analysis has been focused only on the most promising hyperparameters value, fine-tuning them with a grid search strategy.
% Table generated by Excel2LaTeX from sheet 'Sheet2'

 So, for the AMSGrad optimizer we set $\beta_1=0.86$, $\beta_2=0.98$ and $\epsilon = 10^{-9}$. Finally, as previously introduced, with a preliminary procedure, we linearly increased the learning rate of $\eta$ while observing the value of the loss function in order to estimate the initial value of this important hyperparameter. In conclusion, we fed our model with more than 62000 samples for 150 epochs with a batch size of 128 while cyclically varying the learning rate value with a cosine aneling strategy. All tests have been carried out with the TensorFlow framework on a workstation with 64 GB RAM, Intel Core i7-9700K CPU, and an Nvidia 2080 Ti GPU. 

\subsection{Classification}
 Performance of the classifier was evaluated by user’s
accuracy (UA), producer’s accuracy (PA), overall accuracy (OA), and the kappa coefficient (K) shown in the confusion matrix see Table. \ref{tab:tab4}, which is the most common metric that has been used for classification tasks \cite{b10,b47,b103,b104}. Overall accuracy indicates the overall performance of our proposed Pixel R-CNN architecture by calculating a ratio between the correctly classified total number of pixels and total ground truth pixels for all classes. The diagonal elements of the matrix represent the pixels that were classified correctly for each class. Individual class accuracy was calculated by dividing the number of correctly classified pixels in each category by the total number of pixels in the corresponding row called User's accuracy, and columns called Producer's accuracy. PA indicates the probability that a certain crop type on the ground is classified as such. 
\begin{figure*}[htpb]
\centering
 \includegraphics[scale=0.35]{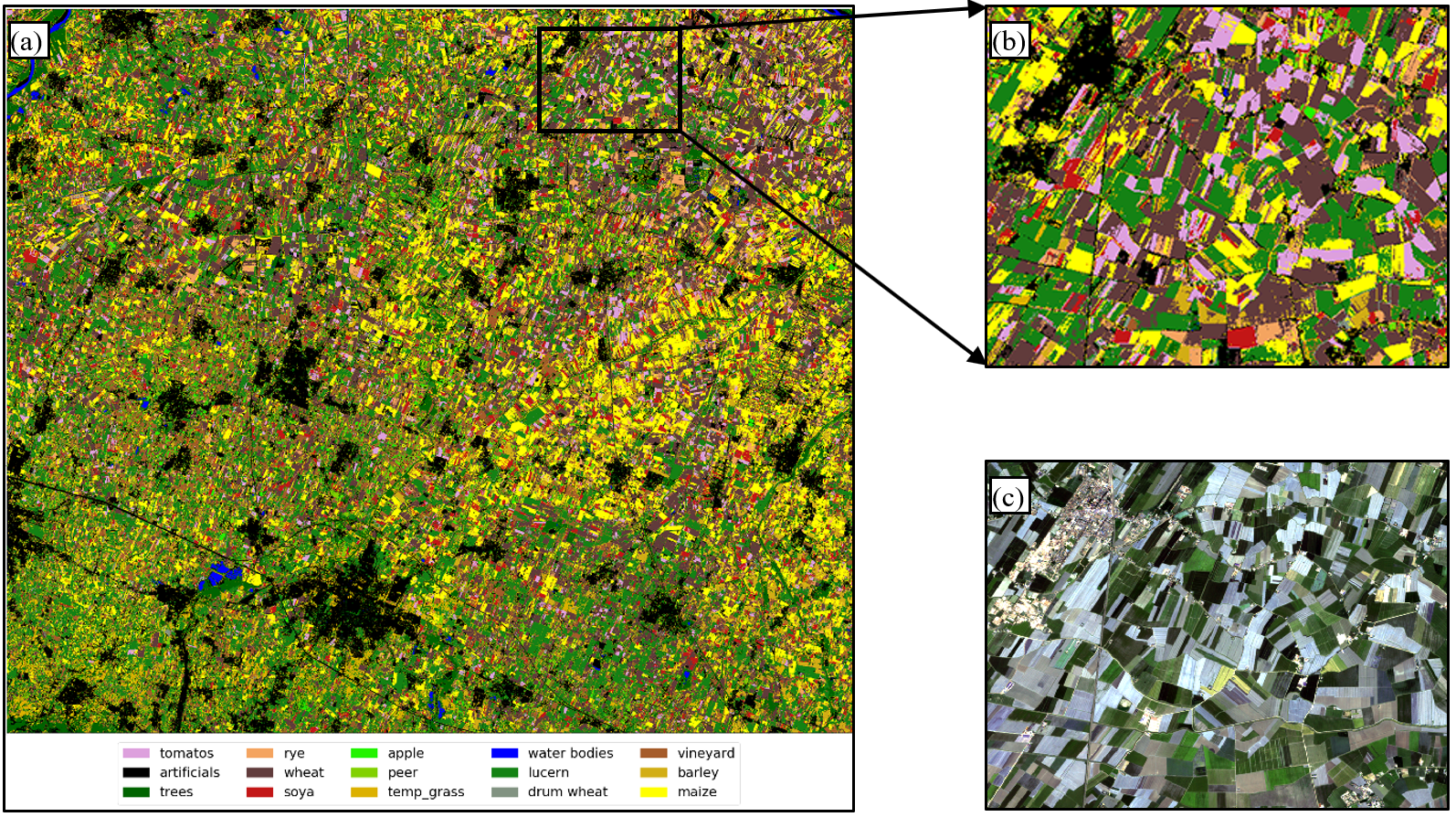}
  \caption {(a). Final classified map using Pixel R-CNN, (b). zoomed in region of the classified map, and (c). Raw Sentinel-2 RGB composite of the zoomed region.}
\label{fig:fig8}
\end{figure*}
UA represents the probability that a pixel classified in a given class belongs to that class.  Our proposed pixel-based Pixel R-CNN method achieved OA=96.5\% and Kappa=0.914 with 15 number of classes for a diverse large scale area, which exhibits significant improvement as compared to other mainstream methods.
Water bodies and trees stand highest in terms of UA with 99.1\% and 99.3\%, respectively. That is mainly due to intra-class variability and the minor change of NIR band reflectance over the time, which was easily learned by our Pixel R-CNN. Most of the classes, including the major types of crops such as Maize, Wheat, Lucerne, Vineyard, Soya, Rye, and Barley, were classified with more than 95\% UA. Grassland being the worst class, which was classified with the PA = 65\% and UA = 63\%. The major confusion of grassland class was with Lucerne and Vineyard.  It is worth mentioning that, Artificial class, which belongs to roads, buildings, urban areas, represents mixed nature of pixel reflectances, was accurately detected with UA = 97\% and PA=99\%.
% Table generated by Excel2LaTeX from sheet 'Sheet1'
\begin{table}
  \scriptsize
  \caption{Obtained confusion matrix.}
     \centering
     \noindent\makebox[\textwidth]{
    \begin{tabular}{p{8.83em}rrrrrrrrrrrrrrrrc}
    \toprule
    \textbf{Ground Truth} & \multicolumn{15}{c}{\textbf{Classified Classes}}                                                           &       &  \\
    \multicolumn{1}{r}{} & \multicolumn{15}{c}{}                                                                                                 & \multicolumn{1}{p{1.715em}}{Total} & \multicolumn{1}{p{1.57em}}{PA} \\
\cmidrule{2-16}    \multicolumn{1}{r}{} & \multicolumn{1}{p{1.855em}}{TM} & \multicolumn{1}{p{2em}}{AR} & \multicolumn{1}{p{2em}}{TR} & \multicolumn{1}{p{1.855em}}{RY} & \multicolumn{1}{p{1.855em}}{WH} & \multicolumn{1}{p{1.855em}}{SY} & \multicolumn{1}{p{1.43em}}{AP} & \multicolumn{1}{p{1.43em}}{PR} & \multicolumn{1}{p{1.57em}}{GL} & \multicolumn{1}{p{1.57em}}{WT} & \multicolumn{1}{p{2em}}{LN} & \multicolumn{1}{p{1.57em}}{DW} & \multicolumn{1}{p{1.855em}}{VY} & \multicolumn{1}{p{1.43em}}{BL} & \multicolumn{1}{p{2em}}{MZ} &       &  \\
    \midrule
    Tomatoes (TM) & \textbf{1096} & 0     & 0     & 0     & 4     & 11    & 0     & 0     & 0     & 0     & 0     & 0     & 0     & 0     & 0     & 1111  & 98\% \\
    Artificial (AR) & 0     & \textbf{3752} & 8     & 1     & 2     & 0     & 2     & 1     & 9     & 9     & 12    & 2     & 6     & 0     & 4     & 3808  & 99\% \\
    Trees (TR) & 0     & 31    & \textbf{2967} & 1     & 0     & 0     & 0     & 3     & 10    & 0     & 17    & 0     & 2     & 0     & 0     & 3031  & 98\% \\
    Rye (RY) & 0     & 1     & 0     & \textbf{1960} & 25    & 0     & 0     & 0     & 0     & 0     & 0     & 0     & 0     & 5     & 0     & 1991  & 98\% \\
    Wheat (WH) & 38    & 7     & 0     & 221   & \textbf{4981} & 6     & 0     & 0     & 10    & 0     & 14    & 1     & 2     & 38    & 42    & 5360  & 93\% \\
    Soya (SY) & 3     & 0     & 0     & 0     & 3     & \textbf{1226} & 0     & 0     & 0     & 0     & 11    & 0     & 3     & 0     & 41    & 1287  & 95\% \\
    Apple (AP) & 0     & 0     & 0     & 0     & 0     & 0     & \textbf{142} & 0     & 0     & 0     & 2     & 0     & 21    & 0     & 0     & 165   & 86\% \\
    Peer (PR) & 0     & 0     & 11    & 0     & 0     & 0     & 27    & \textbf{124} & 0     & 0     & 0     & 0     & 6     & 0     & 0     & 168   & 73\% \\
    Grassland (GL) & 0     & 39    & 3     & 7     & 0     & 1     & 0     & 0     & \textbf{239} & 0     & 72    & 0     & 3     & 0     & 4     & 368   & 65\% \\
    Water (WT) & 0     & 0     & 0     & 0     & 0     & 0     & 0     & 0     & 0     & \textbf{906} & 0     & 0     & 0     & 0     & 0     & 906   & 100\% \\
    Lucerne (LN) & 0     & 0     & 0     & 2     & 0     & 2     & 0     & 0     & 48    & 0     & \textbf{7250} & 0     & 26    & 0     & 10    & 7338  & 98\% \\
    Durum.Wheat (W) & 0     & 4     & 0     & 0     & 0     & 0     & 0     & 0     & 2     & 0     & 0     & \textbf{322} & 0     & 0     & 0     & 328   & 98\% \\
    Vineyard (VY) & 11    & 7     & 4     & 4     & 11    & 1     & 50    & 1     & 21    & 0     & 93    & 0     & \textbf{2139} & 0     & 7     & 2349  & 91\% \\
    Barley (BL) & 0     & 1     & 0     & 2     & 24    & 0     & 0     & 0     & 1     & 0     & 1     & 0     & 0     & \textbf{817} & 0     & 846   & 96\% \\
    Maize (MZ) & 17    & 14    & 0     & 0     & 10    & 24    & 0     & 3     & 10    & 0     & 16    & 1     & 6     & 0     & \textbf{7689} & 7790  & 99\% \\
    Total & 1165  & 3856  & 2993  & 2198  & 5060  & 1271  & 221   & 132   & 350   & 915   & 7488  & 326   & 2214  & 860   & 7797  &       &  \\
    UA    & 94\%  & 97\%  & 99\%  & 89\%  & 98\%  & 96\%  & 64\%  & 93\%  & 68\%  & 99\%  & 96\%  & 99\%  & 96\%  & 95\%  & 98\%  &       &  \\
    \bottomrule
    \end{tabular}}
  \label{tab:tab4}%
\end{table}

\begin{table*}[htb]
  \centering
  \caption{An overview and performance of recent studies.}
  \noindent\makebox[\textwidth]{
    \begin{tabular}{p{16.215em}p{10em}p{9.285em}p{9.57em}ll}
    \toprule
    \textbf{Study} & \multicolumn{5}{p{37.57em}}{\textbf{Details}} \\
\cmidrule{2-6}    \multicolumn{1}{c}{} & \textbf{Sensor} & \textbf{Features} & \textbf{Classifier} & \multicolumn{1}{p{4.785em}}{\textbf{Accuracy}} & \multicolumn{1}{p{3.93em}}{\textbf{ Classes}} \\
    \midrule
    Our   & Sentinel-2 & BOA Reflectances & Pixel R-CNN & 96.50\% & 15 \\
    Ru{\ss}wurm and K\"orner\cite{b102}, 2018 & Sentinel-2 & TOA Reflectances & Recurrent Encoders & 90\%  & 17 \\
    Skakun et al. \cite{b104}, 2016 & Radarsat-2 + Landsat-8 & Optical+SAR & NN and MLPs & 90\%  & 11 \\
    Conrad et al. \cite{b103}, 2014 & RapidEye & Vegetation Indices & RF and OBIA & 86\%  & 9 \\
    Vuolo et al. \cite{b105}, 2018 & Sentinel-2 & Optical & RF    & \multicolumn{1}{p{4.785em}}{91-95\% } & 9 \\
    Hao et al. \cite{b24}, 2015 & MODIS & Stat + phenological & RF    & 89\%  & 6 \\
    J.M. Pena-Barrag{\'a}n \cite{b106}, 2011 & ASTER & Vegetation Indices & OBIA+DT & 79\%  & 13 \\
    \bottomrule
    \end{tabular}}
  \label{tab:tab5}%
\end{table*}%

For class Apple,  obtained PA was 86\% while UA = 68\%, which shows that 86\% of the ground truth pixels were identified as Apple, but only 68\% of the pixels classified as Apple in the classification were actually belonged to class Apple. Some Pixels (see Table. \ref{tab:tab4} ) belongs to Peer and Vineyard were mistakenly classified as Apple. 
 The final classified map is shown in Fig. \ref{fig:fig8} with the example of the zoomed part and actual RGB image. To the best of our knowledge, a multi-temporal benchmark dataset is not available to compare classification approaches on equal footings. There are some data sets available online for crop classification without having ground truth of other land cover types such as Trees, Artificial land (build ups), Water bodies, Grassland. Therefore it is difficult to compare classification approaches on equal footings. Indeed, a direct quantitative comparison of the classification performed in these studies is difficult due to various dependencies such as the number of evaluated ground truth samples, the extent of the considered study area, and the number of classes to be evaluated. Nonetheless, we provided an overview of recent studies and their performances of the study domain by their applied approaches, the number of considered classes, used sensors, and achieved overall accuracy in Table. \ref{tab:tab5}. Hao et al. \cite{b24}, achieved 89\% OAA by using RF classifier on the extracted phenological features from MODIS time-series data. They determined that good classification accuracies can be achieved with handcrafted features and classification algorithms if the temporal resolution of the data is sufficient. Though, the MODIS sensor data is not suitable for classification of the areas of large homogeneous regions due to its low spatial resolution (500m).  Conrad et al. \cite{b102} used high spatial resolution data from the RapidEye sensor and achieved 90\% OAA for nine considered classes. In \cite{b103}, features from optical and SAR were extracted and used by the committee of neural networks of multilayer perceptrons to classify a diverse agriculture region considerably. Recurrent encoders have been employed in \cite{b101} to classify a large area for 17 considered classes using high spatial resolution (10m) sentinel-2 data and achieved 90\% OAA,which proved that recurrent encoders are useful to capture the temporal information of spectral features that leads to higher accuracy. Voulo et al. \cite{b104} also used sentinel-2 data and achieved a maximum 95\% classification accuracy using RF classifier but nine classes were considered.
  \begin{figure}[t]
\centering
 \includegraphics[scale=0.7]{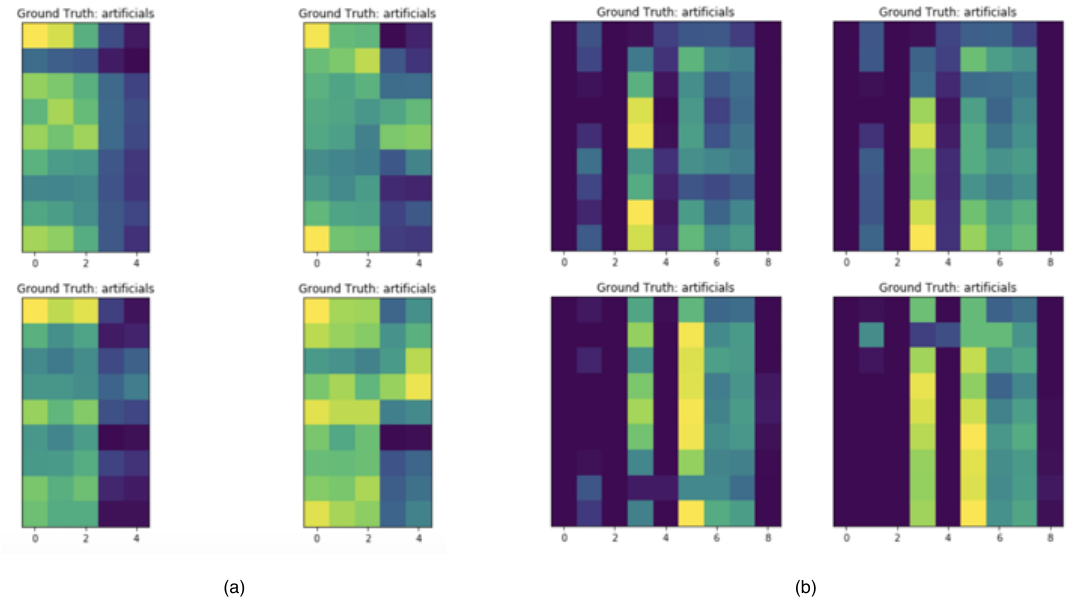}
  \caption {Visual representation of the activation of the internal neurons of Pixel R-CNN, where darker color are values close to zero and vice versa. (a). four samples of the same class "artificials", (b). related activations inside the network at the output of the TimeDistributedDense layer $Y_{(timeD)}$. It is interesting to notice how the four inputs are pretty different from each other, but the network representations already at this level are similar.}
\label{fig:figxi}
\end{figure}
 
 In conclusion, it is interesting to notice neuron activation inside the network during the classification process. Indeed, it is possible to plot unit values when the network receives specific inputs and compare how model behaviors change. In Fig. \ref{fig:figxi} four samples, belonging to the same class "artificials," feed the model creating individual activations in the input layer. Even if they all belong to the same class, the four instances took into account present a noticeable variance. Either the spectral features in a specific time instance (rows) or their temporal variation (columns) present different patterns that make them difficult to classify. However, already after the first LSTM layer with the TimeDistributedDense block, the resulting 9x9 matrices $Y_{(timeD)}$ have a clear pattern that can be used by the following layers to classify the different instances in their respective classes. So, the network during the training process learns to identify specific temporal schemes, that allows making strong distributed and disentangle representations.

\subsection{Non deep learning classifiers}
We tried four other traditional classifiers on the same dataset for comparison, which are Support Vector Machine (SVM), Kernal SVM, Random Forest (RF), and XGBoost. These are well-known classifiers for their high performances and also considered as baseline models in classification tasks \cite{b97}. SVM can perform nonlinear classification using kernel functions by separating hyperplanes. A widely used RF classifier is an ensemble of decision trees based on the bagging approach \cite{b98,b99}. XGBoost is state of the art classifier based on gradient boosting model of decision trees, which attracted much attention in the machine learning community. RF and SVM have been widely used in remote sensing applications \cite{b23,b24,b40}. Each classifier involves hyperparameters that need to be tuned at the time of classification model development.
% Table generated by Excel2LaTeX from sheet 'Sheet1'
\begin{table}[htpb]
      \centering
  \caption{Comparison of Pixel R-CNN with non-deep learning classifiers.}
    \begin{tabular}{p{6.145em}p{14.5em}l}
    \toprule
    \textbf{Model} & \textbf{Parameters} & \multicolumn{1}{p{3.215em}}{\textbf{OA}} \\
    \midrule
    SVM & C: 0.01, 0.1, \textbf{1}, 10, 100, 1000  & 79.50\% \\
    \multicolumn{1}{r}{} & Kernel: linear &  \\
    \midrule
    Kernel SVM & C: 0.01, 0.1, \textbf{1}, 10, 100, 1000   & 76.20\% \\
    \multicolumn{1}{r}{} & Kernel: \textbf{rbf}   &  \\
    \multicolumn{1}{r}{} & Gamma: \textbf{0.1}, 0.2, 0.3, 0.4, 0.5, 0.6, 0.7, 0.8 &  \\
    \midrule
    Random Forest & n\_estimators: 10, 20, 100, 200, \textbf{500}    max\_depth: \textbf{5}, 10, 15, 30      min\_samples\_split: 3, \textbf{5}, 10, 15, 30 min\_samples\_leaf: 1, 3, \textbf{5}, 10 & 77.90\% \\
    \midrule
    XGBoost & learning\_rate: \textbf{0.01}, 0.02, 0.05, 0.1 & 77.60\% \\
    \multicolumn{1}{r}{} &  gamma: 0.05, \textbf{0.1}, 0.5, 1  &  \\
    \multicolumn{1}{r}{} & max\_depth: 3, \textbf{7}, 9, 20, 25       min\_child\_weight: 1, 5, 7, \textbf{9 }&  \\
    \multicolumn{1}{r}{} &  subsamples: 0.5, \textbf{0.7}, 1  &  \\
    \multicolumn{1}{r}{} & colsample\_bytree: \textbf{0.5}, 0.7, 1 &  \\
    \multicolumn{1}{r}{} & reg\_labda:  0.01, 0.1, \textbf{1}  &  \\
    \multicolumn{1}{r}{} & reg\_alpha: 0, 0.1, 0.5, \textbf{1} &  \\
    \midrule \textbf{Pixel R-CNN} & Mentioned in experimental settings & \textbf{96.50\%} \\
    \bottomrule
    \end{tabular}%
  \label{tab:tab6}%
\end{table}%

We followed "random search" approach to optimize major hyperparameters \cite{b100}. Best values of hyperparameters were selected based on classification accuracy achieved for the validation set, and are highlighted with bold letters in Table. \ref{tab:tab6}. Further details about hyper parameters and achieved overall accuracy (OA) for SVM, Kernal SVM, RF, and XGBoost are reported in Table. \ref{tab:tab6} From these non-deep learning classifiers, SVM stands highest with OA = 79.6\% while RF , Kernel SVM, and XGBoost achieved 77.5\%, 76.8\% and 77.2\% respectively. From the results presented in Table. \ref{tab:tab6}, our proposed Pixel R-CNN based classifier achieved OA = 96.5\%, which is far better results than the non deep learning classifiers. Learning temporal and spectral correlations from multi-temporal images considering large data set is challenging for traditional non-deep learning techniques. The introduction of deep learning models in the remote sensing domain brought more flexibility to exploit temporal features in such a way that it can increase the amount of information to gain much better and reliable results for classification tasks. 

%%%%%%%%%%%%%%%%%%%%%%%%%%%%%%%%%%%%%%%
\section{Conclusion}
In this study, we developed a novel deep learning model with Recurrent and Convolutional Neural Network called Pixel R-CNN to perform Land Cover and Crop Classification by using multitemporal decametric sentinel-2 imagery of central north part of Italy. Our proposed Pixel R-CNN based architecture exhibits significant improvement as compared to other mainstream methods by achieving 96.5\% overall accuracy with kappa=0.914 for 15 number of classes. We also tested widely used non-deep learning classifiers such as SVM, RF, SVM kernel, and XGBoost to compare with our proposed classifier and revealed that these methods are less effective, especially when the temporal features extraction is the key to increase classification accuracy. The main advantage of our architecture is the capability of automated features extraction by learning time correlation of multiple images, which reduces manual feature engineering and modeling crops phenological stages.

%%%%%%%%%%%%%%%%%%%%%%%%%%%%%%%%%%%%%%%%%%
\vspace{6pt} 

%%%%%%%%%%%%%%%%%%%%%%%%%%%%%%%%%%%%%%%%%%
%% optional
%\supplementary{The following are available online at \linksupplementary{s1}, Figure S1: title, Table S1: title, Video S1: title.}

% Only for the journal Methods and Protocols:
% If you wish to submit a video article, please do so with any other supplementary material.
% \supplementary{The following are available at \linksupplementary{s1}, Figure S1: title, Table S1: title, Video S1: title. A supporting video article is available at doi: link.}

%%%%%%%%%%%%%%%%%%%%%%%%%%%%%%%%%%%%%%%%%%
\section*{Acknowledgments}
This work has been developed with the contribution of the Politecnico di Torino Interdepartmental Centre for Service Robotics PIC4SeR (https://pic4ser.polito.it) and SmartData@Polito (https://smartdata.polito.it).

\bibliographystyle{unsrt}  
%\bibliography{references}  %%% Remove comment to use the external .bib file (using bibtex).
%%% and comment out the ``thebibliography'' section.

%%% Comment out this section when you \bibliography{references} is enabled.

\end{document}